\documentclass{article}

\usepackage[preprint]{corl_2024} 
\usepackage{tikz}
\usetikzlibrary{shapes, arrows, fit, calc, positioning, automata, backgrounds, patterns, decorations.pathmorphing, decorations.markings}
\usepackage[thinc]{esdiff}
\graphicspath{{figs/}}
\usepackage{xcolor}
\usepackage{graphicx, amssymb, array, rotating}
\usepackage[utf8]{inputenc}
\usepackage{svg}
\usepackage{siunitx}
\usepackage{booktabs}
\usepackage{xfrac}
\usepackage[draft, markup=bfit, deletedmarkup=sout, authormarkup=brackets]{changes}
\usepackage[thinc]{esdiff}
\usepackage{times}
\usepackage{multicol}
\usepackage[bookmarks=true]{}
\usepackage{soul}
\usepackage{color}
\usepackage{graphics}
\usepackage{epsfig} 
\usepackage{caption} 
\usepackage{mdwmath}
\usepackage{mdwtab}
\usepackage{url}
\usepackage{float}
\usepackage[makeroom]{cancel}
\usepackage{amsmath,bm}
\usepackage{dblfloatfix}
\usepackage{lipsum}
\usepackage{relsize}
\usepackage{algpseudocode}
\usepackage[ruled,vlined]{algorithm2e}
\usepackage{xspace}
\usepackage{mathtools}
\usepackage{amsfonts}
\usepackage{cite}
\usepackage{flushend}
\usepackage[normalem]{ulem} 
\usepackage{marginnote}
\usepackage{atbegshi}
\usepackage{subfig} 
\usepackage{threeparttable}
\usepackage[font={small}]{caption}
\usepackage[draft, markup=bfit, deletedmarkup=sout, authormarkup=brackets]{changes}
\title{Tactile-Morph Skills: Energy-Based Control Meets Data-Driven Learning}

\author{
  Anran Zhang, Kübra Karacan, Hamid Sadeghian, Yansong Wu, Fan Wu, and Sami Haddadin \\
 Munich Institute of Robotics and Machine Intelligence \\ Technical University of Munich
\\  \{name.lastname@tum.de\}\\
}

\begin{document}

\def\baselinestretch{0.975}\selectfont

\maketitle


\begin{abstract}
Robotic manipulation is essential for modernizing factories and automating industrial tasks like polishing, which require advanced tactile abilities. These robots must be easily set up, safely work with humans, learn tasks autonomously, and transfer skills to similar tasks. Addressing these needs, we introduce the tactile-morph skill framework, which integrates unified force-impedance control with data-driven learning. Our system adjusts robot movements and force application based on estimated energy levels for the desired trajectory and force profile, ensuring safety by stopping if energy allocated for the control runs out. Using a Temporal Convolutional Network, we estimate the energy distribution for a given motion and force profile, enabling skill transfer across different tasks and surfaces. Our approach maintains stability and performance even on unfamiliar geometries with similar friction characteristics, demonstrating improved accuracy, zero-shot transferable performance, and enhanced safety in real-world scenarios. This framework promises to enhance robotic capabilities in industrial settings, making intelligent robots more accessible and valuable.
\end{abstract}



\section{Introduction} \label{sec:intro}
Robot manipulation is central to achieving the promise of robotics. In recent years, learning-based approaches have injected substantial growth in research on the problem of robot manipulation. However,~\emph{contact-rich tactile manipulation} is still considered problematic and tedious to manage due to the inherent complexity of contact behavior when the environment model is uncertain \citep{Suomalainen2022}. Besides, even state-of-the-art Robotic Simulator fails to capture this dynamic. Meanwhile, this aspect is crucial for modernizing traditional factories and automating complex tasks like polishing and wiping. These contact-rich tasks require advanced tactile abilities for robots to adapt to various environments and conditions. To be beneficial in industrial settings, robots must: (i) be easy to set up and safely work with humans, (ii) learn tasks autonomously, and (iii) transfer skills to similar tasks \citep{haddadin_tactile_internet_2019}. Developing such solutions could make intelligent robots more accessible and valuable, especially as we face challenges like a shortage of skilled workers and demographic changes. However, contact-rich tasks are challenging to automate and require learning, robotics, and control expertise.

Energy-based control is a feasible solution to handle contact-rich tasks, which ensures robots handle tasks compliantly, safely, and efficiently in various environments. This approach is particularly relevant to our proposed solution, as it helps to understand the energy needed to interact with the world and stabilize the robot's actions. However, energy-based approaches may decline performance when the energy amount is not configured appropriately and fails to complete the task. For instance, polishing a scratch at different locations on a car door would require particular homogeneous force exertion while moving in different trajectories. Each trajectory for the same surface would require a different energy value. Thus, learning the correct energy levels for tasks and understanding the performance of the energy needed to interact with the world helps stabilize the robot's actions. 

To address these needs, we present a novel approach—the Tactile-Morph Skills framework—which merges unified force-impedance control with data-driven learning. First, by adjusting tactile behavior based on energy levels—what we call 'tactile-morph '—our system modifies the robot's movements and force application, stopping if it runs out of energy to ensure safety and efficiency. Second, to estimate the energy distribution of the surface for a given motion and force profile, we input time-variant sequential motion and time-invariant force policy to the Temporal Convolutional Network. This network, a type of neural network designed for processing time series data, serves as the backbone of our proposed solution by helping to estimate the energy distribution of the surface accurately. Third, evaluating our framework's efficacy under real-world scenarios, we compare the accuracy of energy estimation with expert estimation and baseline models, demonstrating the zero-shot transferable performance on different surfaces and safety performance on real robots. To the best of the authors' knowledge, our method is the first to describe an energy distribution for contact-rich tactile skills safely and successfully while applying force. Our approach ensures the robot's stability and performance while following the desired motion and force profile, even in unfamiliar geometries with similar friction characteristics. This innovation enables the transfer of contact skills between different trajectories and force profiles on the surface material with varying geometries, providing a robust framework for autonomous and adaptive robotic manipulation. Our contributions: 
\begin{itemize}
  \item[I] \hangindent=2em Introducing a data-driven method for estimating contact-rich task energy distribution. Our method achieved accuracy over other state-of-the-art methods.
  \item[II] \hangindent=2em Integrating energy distribution modeling with unified force-impedance control within the Tactile-Morph Skills Framework, with validated safety performance in real robotic experiments.
  \item[III] \hangindent=2em Release of publicly available code and data for the Tactile-Morph Skills Framework.
\end{itemize}

\section{Related Literature} \label{sec:sota}

\subsection{Robotic Tactile Skills}
\textbf{Robotic tactile skills}, such as grinding and wiping, require precise control of interactions at the end-effector. These skills necessitate a motion generation unit coupled with a force policy. The literature presents various perspectives on combining motion generation and force control in robotic manipulators. For instance, Zielinski et al.~\citep{Zielinski2010} classify manipulator behavior into three phases: free-motion (where force is insignificant), exerting generalized forces, and transitions between these behaviors. Developing complex perceptuomotor skills is crucial as robots encounter real-world challenges~\citep{Pastor2011}. 

\textbf{Robot Tactile Skills in Simulation}. While progress has been made in simulating and transferring manipulation policies to robots, current efforts primarily address tasks with limited contact richness~\citep{Todorov2012, fu2016, narang2022, Radosavovic2023}. However, due to contact simulation fidelity and collision modeling challenges for articulated rigid-body systems, simulations may fail when the task policy is learned to be performed in highly unconstrained environments such as manufacturing. Thus, contact-rich skills should be designed as flexible and adaptive as possible.

\subsection{Control Techniques in Contact-rich Skills}
\textbf{Position control} has been proven impractical for defining tactile skills, necessitating alternative control methods~\citep{Hogan1984}. \textbf{Force control} has been extensively explored, with controllers validated based on constant force values or thresholds/constraints~\citep{Kulakov2015, Ficuiello2015, Ott2015, He2016, Cherubini2016}. However, robots must be robust enough to operate autonomously in unstructured environments and perform tasks despite perception uncertainties~\citep{Pastor2012, Kim2021, Wang2022, Liang2022, Zhang2024}. Wavelet neural networks have been effectively utilized for compliant force tracking in environments with unknown geometry~\citep{Hamedani2021}.

\textbf{Impedance control} is a prominent approach that imposes a dynamic behavior between external interaction and desired motion, rather than independently tracking motion or force trajectories~\citep{Hogan1984}. These dynamics can be applied in a robot manipulator's joint, operational, or redundant spaces~\citep{sadeghian2013task}. Typically, impedance parameters such as inertia, damping, and stiffness are constant in each direction based on the assigned tasks, providing a passive mapping between external force inputs and the robot's motion outputs. This ensures the system does not generate additional energy during passive environment interactions, maintaining stability~\citep{ott2008cartesian}. However, adapting impedance parameters can offer more human-like behavior during external interactions~\citep{yang2011human, Luo2019, Karacan2022}. Despite its benefits, varying impedance parameters can compromise control loop passivity, leading to instability. To address this, energy tanks adjust the closed-loop system dynamics and ensure passivity~\citep{Ferraguti2013,haddadin2024unified}.

\textbf{Virtual energy tanks} are integrated to identify potential instabilities from stiffness variations and force regulations to ensure stability even amid dynamic changes~\citep{Stramigioli2015}. However, precisely initializing the energy tank for specific tasks has been challenging. Since proving stability does not guarantee successful task completion~\citep{Shahriari2020RAL}. Insufficient tank energy can deactivate the force controller or transition impedance control to compliance control, preventing correct task execution. To solve this, initial task energy is the minimum energy required in the tanks to meet skill requirements. Knowing this task energy or its lower bound before execution ensures stability and performance. The most straightforward strategy involves using sufficiently high energy. Still, for improved safety and process monitoring, leveraging model-based~\citep{fu2016}, data-driven~\citep{Zanella2024}, or model-informed hybrid approaches~\citep{Bogdanovic2020, Khader2021} is beneficial.

\section{Methodology}\label{sec:method}
\begin{figure}
    \centering
    \includegraphics[width=\linewidth]{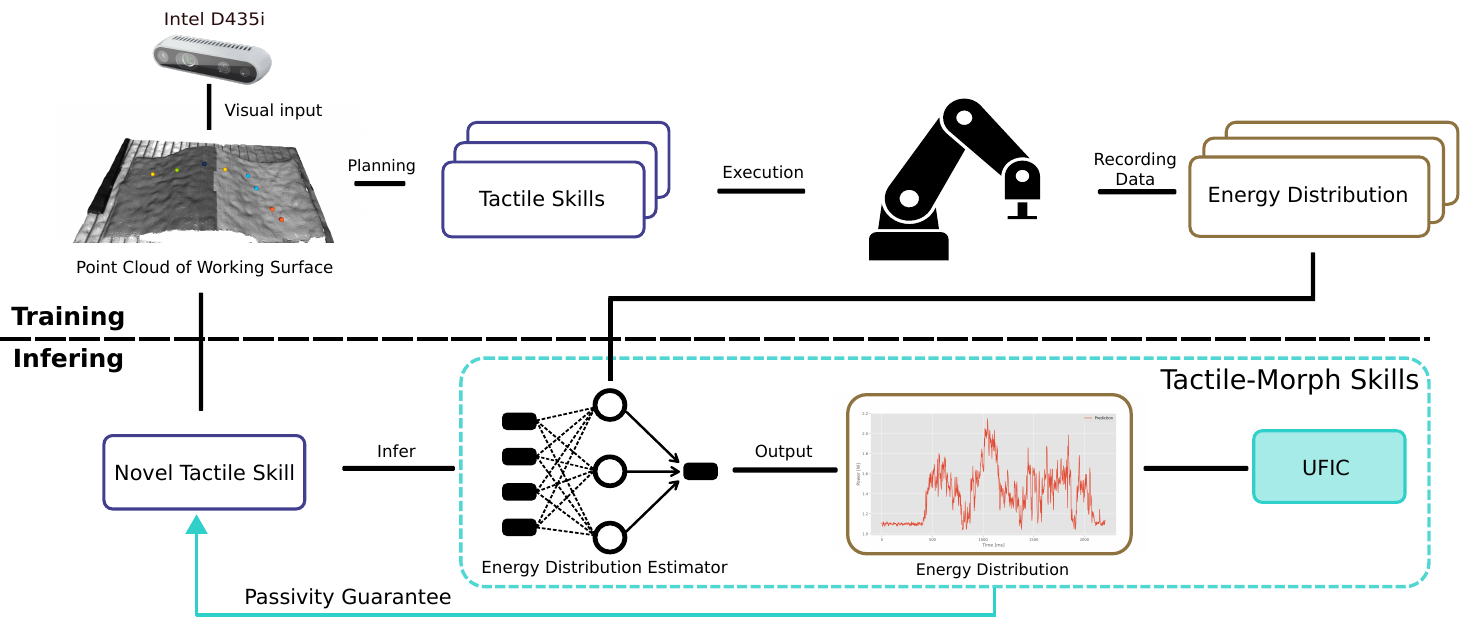}
    \caption{\textbf{Pipeline Overview}: Tactile-Morph Skills Framework is composed of an energy distribution estimator and unified force-impedance control(UFIC). During the training phase, a set of tactile skills, designed based on basic motion patterns, is planned using point cloud observations of the working surface. These tactile skills are executed by the robot, and the corresponding energy distribution is recorded in real-time. The collected data is then utilized to train our energy distribution estimation model. Upon completion of the training process, the energy distribution for any novel tactile skill sampled from the working surface can be accurately estimated. By integrating the estimated energy distribution with UFIC, we can provide a formal guarantee of passivity to any contact-rich tactile skill.}
    \label{fig:workflow}
\end{figure}

The Tactile-Morph Skills framework is designed to achieve three primary objectives: enabling a system to i.) "understand" its environment, adapt the prior policy, and learn the correct control for varying environmental conditions, and ii.) robustly and safely adjust to unforeseen process conditions, as illustrated in Fig.~\ref{fig:workflow}. To achieve these goals, we integrated a Temporal Convolutional Network (TCN) to estimate the energy distribution of manipulation tasks, allowing precise management of the energy tank status. In the following sections, we introduce the details of the nominal skill, the TCN-based energy status estimation, and the overall structure of our proposed Tactile-Morph skills framework.

\subsection{Derivation of Nominal Tactile Skill \label{nominal_skill}}

For a robot manipulator under unified force-impedance control during contact with gravity compensation, the Lagrangian dynamics in Cartesian space are expressed as:
\begin{align}
\bm{M}_\mathrm{C}\ddot{\bm{x}} + \bm{C}_\mathrm{C}\dot{\bm{x}}+ \bm{D}_\mathrm{C}\dot{\bm{x}} +\bm{f}_\mathrm{g}&= \bm{f}_\mathrm{robot}+\bm{f}_\mathrm{ext}\,.
\end{align}
Here, $\bm{f}_\mathrm{ext} \in \mathbb{R}^6$ denotes the external wrench applied to the base frame. The robot mass matrix is represented by $\bm{M}_\mathrm{C}(\bm{q})$, where $\bm{q}$ is the joint configuration. The Coriolis and centrifugal effects are captured by $\bm{C}_\mathrm{C}(\bm{q}, \dot{\bm{q}}) \in \mathbb{R}^\mathrm{6 \times 6}$, and $\bm{f}_\mathrm{g}$ represents the gravity vector in Cartesian space. Additionally, $\bm{f}_\mathrm{robot}$ is the wrench applied by the robot. Compliance control, a subset of impedance control, omits inertia shaping and consequently excludes feedback of external forces. The dissipative term is $\bm{D}_\mathrm{C} \in \mathbb{R}^\mathrm{6\times6}$, $\bm{x} \in \mathbb{R}^6$ denotes the current pose of the end-effector, and $\tilde{\bm{x}}$ represents the pose error. The force applied by the robot $\bm{f}_\mathrm{robot}$, including the control force $\bm{f}_\mathrm{cntr} \in \mathbb{R}^6$ for tracking desired motion $\bm{x}_\mathrm{d} \in \mathbb{R}^6$ by impedance control $\bm{f}_\mathrm{imp}\in \mathbb{R}^6$ and the desired force policy $\bm{f}_\mathrm{d} \in \mathbb{R}^6$ by force control $\bm{f}_\mathrm{frc}\in \mathbb{R}^6$, is given by:
\begin{align}
     \bm{f}_\mathrm{robot} &= \bm{f}_\mathrm{cntr}+\bm{f}_\mathrm{g}\,, 
     \\
     \bm{f}_\mathrm{cntr}&= \bm{f}_\mathrm{imp} + \bm{f}_\mathrm{frc}\,.
\end{align}
To analyze the passivity of the controller, the storage function $E_\mathrm{robot}$ for the Cartesian robot dynamics, representing the robot's kinetic energy, is:
\begin{align}
    E_\mathrm{robot} &= \frac{1}{2}\dot{\bm{x}}^T\bm{M}_\mathrm{C}\dot{\bm{x}}\,,
\end{align}
with its time derivative $\dot{E}_\mathrm{robot}$ given by:
\begin{align}
    \dot{E}_\mathrm{robot} &= \frac{1}{2}\dot{\bm{x}}^T\dot{\bm{M}}_\mathrm{C}\dot{\bm{x}}+\dot{\bm{x}}^T\bm{M}_\mathrm{C}\ddot{\bm{x}} \,,
    \\
    &=\dot{\bm{x}}^T\left(\frac{1}{2}\dot{\bm{M}}_\mathrm{C}\dot{\bm{x}} +\bm{f}_\mathrm{robot}+\bm{f}_\mathrm{ext}- \bm{C}_\mathrm{C}\dot{\bm{x}}-\bm{D}_\mathrm{C}\dot{\bm{x}}-\bm{f}_\mathrm{g} \right)\,,
    \\
    &=\dot{\bm{x}}^T\left(\bm{f}_\mathrm{cntr}+\bm{f}_\mathrm{ext}-\bm{D}_\mathrm{C}\dot{\bm{x}}\right)\,.
\end{align}
To ensure passivity, the robot's power must be dissipated such that:
\begin{align}
     \dot{E}_\mathrm{robot} &=\dot{\bm{x}}^T\bm{f}_\mathrm{cntr}+\dot{\bm{x}}^T\bm{f}_\mathrm{ext}-\dot{\bm{x}}^T\bm{D}_\mathrm{C}\dot{\bm{x}} \leq 0\,.
\end{align}
Due to the uncertainty in the sign of the pair ($\bm{f}_\mathrm{cntr},\dot{\bm{x}}$), the force-impedance controller must be modified to ensure stability by augmenting a virtual energy tank for power transmission between the tank and the impedance controller, with the valve $\sigma(E_\mathrm{tank})$ defined as:
\begin{align}
       \sigma(E_\mathrm{tank})&=  \begin{cases} 1, & E_\mathrm{tank}\geq\epsilon\,,
    \\
    0, & else
    \end{cases}
\end{align}
where, after completing the task at $t=t_\mathrm{final}$, the tank should end up with $\epsilon$ amount of energy:
\begin{align}
    \label{eq:energytank}
    E_\mathrm{tank}(t_\mathrm{final})&=  E_\mathrm{tank}(0) - \int_{0}^{t_\mathrm{final}} \dot{\bm{x}}^T\bm{f}_\mathrm{cntr}\,d\tau\, = \epsilon\,.
\end{align}
This indicates that the controller can control force and motion if the tank has enough energy during the task or deactivate the controller if instability arises. The stable controller is thus:
\begin{align}
 \bm{f}_\mathrm{robot} &= \sigma\bm{f}_\mathrm{cntr}+\bm{f}_\mathrm{g}\,.
\end{align}
Setting the initial energy $E_\mathrm{tank}(0)$ to a constant to fulfill the desired task leads to fine-tuning for each task and corresponding working surface material, geometry, etc. Alternatively, setting it to a large default value may result in unnecessary energy waste. We propose learning the initial tank energy or task energy to overcome this issue. Next, we develop a learning pipeline to precisely manage the energy tank status:
\begin{align}
    E_\mathrm{tank}(0) &= \int_{0}^{t_\mathrm{final}} \dot{\bm{x}}^T\bm{f}_\mathrm{cntr}\,d\tau+\epsilon\,.  
\end{align}

\subsection{Tactile-Morph Skills}

The Tactile-Morph Skills framework regulates energy consumption in \SI{}{J/mm} throughout the task to ensure safety and success. Although the optimal energy budget for the nominal tactile skill can be calculated as in Eq.~\eqref{eq:energytank}, this assumes perfect conditions where the controller precisely follows the desired trajectory and force policy. Environmental uncertainties shift this optimal energy budget. As demonstrated in our experiments, the scalar energy budget is often insufficient, causing failures early in the task. Despite this, the nominal skill provides foundational knowledge for morphable skills, and the task energy trajectory can govern the morph level. The optimal energy trajectory can be estimated as a safety guarantee, regardless of environmental uncertainties and disturbances. 

Our estimation pipeline to determine the energy distribution of a surface $[E_\mathrm{tank}]$, with continuous contact with the working surface, is formulated as a power function $\dot{E}_\mathrm{tank}$ depending on the desired motion $\bm{x}_\mathrm{d}$ and force policy $\bm{f}_\mathrm{d} \in \mathbb{R}^6$. The energy distribution to complete the task, $[E_\mathrm{tank}]$ is computed by integrating the power function over the time interval $t_\mathrm{final}$:
\begin{align}
    \label{eq:power_estimation}
    \dot{\bm{E}}_\mathrm{tank} &= f(\bm{x}_\mathrm{d},\bm{f}_\mathrm{d})\,, \\
    [E_\mathrm{tank}] &=  \int_{0}^{t_\mathrm{final}} \dot{\bm{E}}_\mathrm{tank}\,d\tau\,, \\
    \nu_\mathrm{~input} &=  \nu_{\mathrm{~variant}} \oplus \nu_\mathrm{~invariant}\,.
\end{align}

\begin{figure}
    \centering
    \includegraphics[width=12cm]{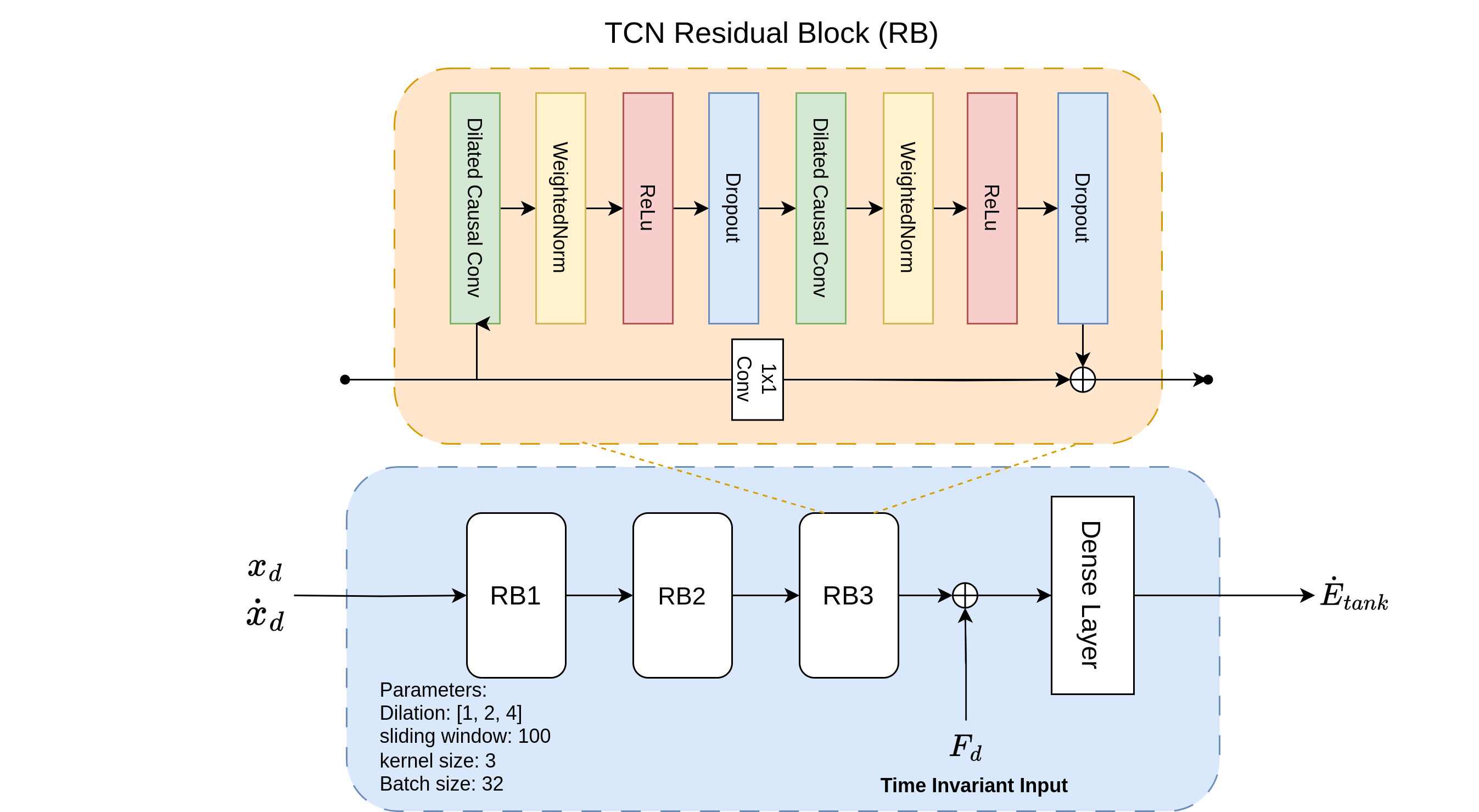}
    \caption{\textbf{Energy Distribution Estimation Pipeline.}}
    \label{fig:tcn}
\end{figure}

This energy consumption function, a continuous function over trajectory and motion, can be learned in a data-driven fashion. We choose a Temporal Convolutional Network (TCN) as the backbone of our pipeline to accommodate sequential input with variable length and enable extended memory, flexibility in the receptive field, and superior performance over LSTM networks~\citep{bai2018}, as shown in Fig.~\ref{fig:tcn}. Our pipeline consists of three TCN modules followed by a two-layer Multilayer Perceptron (MLP) for decoding the energy function, with each TCN block constructed as described in~\citep{KerasTCN}.

\subsection{Pipeline Overview}

\textbf{Data Collection}
We pre-plan the trajectories by leveraging the captured 3D point cloud. The robot then collects the data and executes the trajectories, interacting with the working surface.

\textbf{Pipeline Architecture}
The input of our pipeline is a sequence of time-variant data $\nu_\mathrm{variant}$ and time-invariant force policy $\nu_\mathrm{invariant} = \bm{f}_\mathrm{d}$. The TCN processes the time-variant input $\nu_\mathrm{variant} = \{\bm{x}_\mathrm{d}, \dot{\bm{x}}_\mathrm{d}\}$. The desired velocity profile and the time-invariant input $\nu_\mathrm{invariant}$ are concatenated to ensure enough energy for full-procedure safety. Additionally, learning techniques adjust the morph level of the force-motion policy based on scene geometry encountered during tasks, dynamically adapting force and motion parameters to optimize performance in diverse environments.

\begin{figure}[h]
    \centering
    \begin{minipage}{0.39\textwidth}
        Using the trained model, we estimate the energy needed for any randomly generated tactile skill starting at a specific position on the working surface. An energy heat map is constructed by leveraging the outputs of energy estimation and linear interpolation, as shown in Fig.~\ref{fig:heatmap}. This energy heat map enables fast queries of the required energy trajectory for any tactile skill on the working surface, allowing for robust recovery, extension, and re-planning in real-time.
    \end{minipage}
    \begin{minipage}{0.55\textwidth}
        \centering
        \includegraphics[width=\linewidth]{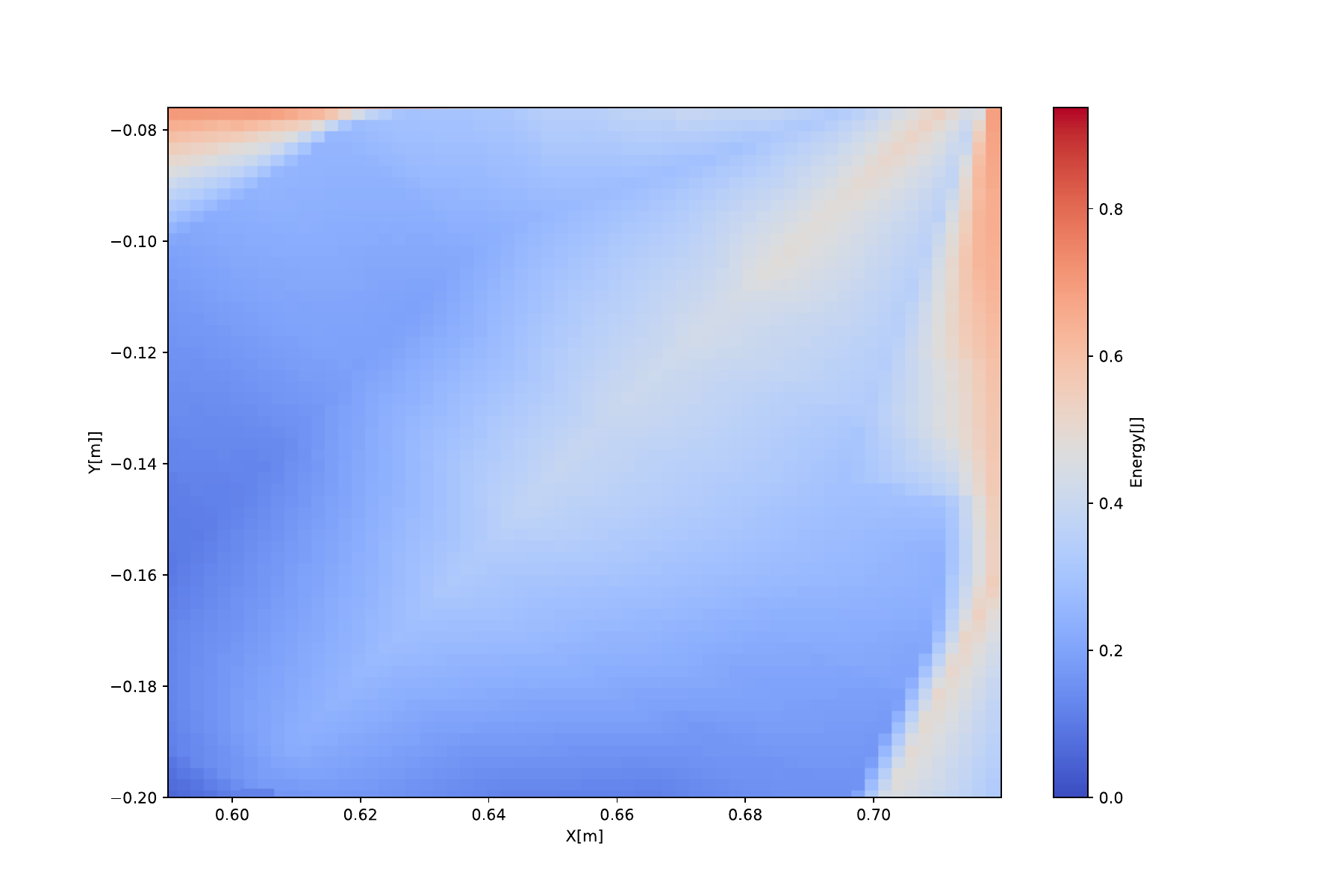}
        \caption{\textbf{Energy Heat Map:} inferring randomly generated trajectories and integrating the power distribution.}
        \label{fig:heatmap}
    \end{minipage}\hfill
\end{figure}
\section{Experimental Validation} \label{sec:exp}
To evaluate the efficacy of the Tactile-morph Skills framework, we systematically identify and formalize performance metrics tailored to address variations in surface geometry and material properties in the specified processes. We evaluate our Tactile-Morph Skills in three scenarios: i) compare the accuracy of energy estimation with educated estimation. ii) demonstrating the zero-shot transferable performance on different surfaces. iii) safety performance on real robots.
\subsection{Performance Evaluation}
To evaluate the performance of our pipeline, we use the mean square error (MSE) and the mean absolute percentage error (MAPE) as performance metrics for the estimated power trajectory and integrated energy sum. Furthermore, the mean percentage error is the most direct reflection of accuracy. We estimate the total energy value $\bm{E}_\mathrm{tank,est}$ instead of distribution based on Eq.~\eqref{eq:dummyestimation}, where $\mu$ is the friction of the working surface and $\bm{f}_{robot}$ and $\dot{\bm{x}}_d$ is the pre-planned force and desired motion velocity. 
\begin{align}
    \label{eq:dummyestimation}
    \bm{E}_\mathrm{tank,est} &= \int_{0}^{t_\mathrm{final}}\mu\dot{\bm{x}}^T_d\bm{f}_{cntr}d\tau\,.
\end{align}
We run performance evaluations over a randomly selected test set. The mean percentage error of estimated power trajectory across unknown tactile skills is around 8\%, demonstrating valid accuracy in estimating power flow over the trajectory. It's worth mentioning that the MAPE of energy sum is even lower, around only 2.28\%. The worst-case estimation can be seen in Fig.~\ref{fig:three_images}a), showing the complexity of a contact-rich scenario that the output power can rapidly evolve in rare cases and that our model failed to follow this dynamic.
\begin{table}[h]
\footnotesize
\begin{threeparttable}
\setlength\tabcolsep{1.5pt} 
\begin{tabular*}{\columnwidth}{@{\extracolsep{\fill}} l cccc}
\toprule
     Methods & MSE & MAPE(\%) & $\mathrm{MAPE}_{sum}(\%)$   \\
\midrule
     Expert's estimation & - & - & $116.193\pm 140.03$\\
     Our pipeline & $0.0434 \pm 0.012$ & $7.931\pm2.23$ & $\bm{2.280}\pm1.84$ \\
\bottomrule
\end{tabular*}
\caption{\textbf{Performance Evaluation on Test Set} Here, we compare the accuracy of the power estimation pipeline with the expert's estimation. The mean percentage error(MAPE) along the whole power trajectory is 11.37\% with a standard deviation of 6.87\%. An expert's estimation can only have total energy value instead of power trajectory and fails to achieve satisfying results. This requires trial and error in experiments.}
\label{tab:Performance}
\end{threeparttable}
\end{table}
\vspace{-18pt}

\subsection{Transferable Tests on Different Surface Geometries} \label{transferable exp}
This section demonstrates that our method can understand the geometry information embedded in the pre-designed trajectory and estimate power distribution across various domains. 

\begin{figure}[h]
    \centering
    \includegraphics[width=\linewidth]{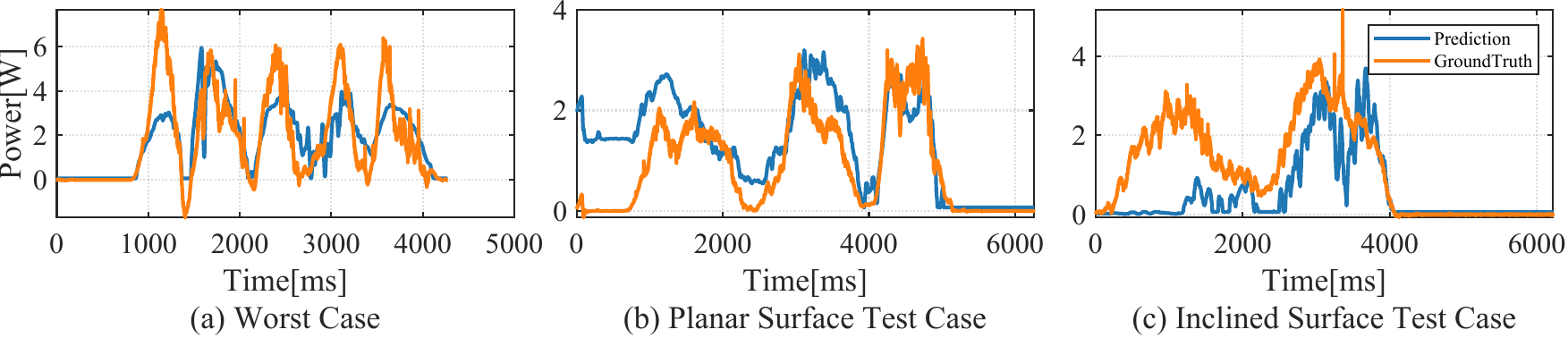} 
    \caption{\textbf{Estimated Power Trajectory} in comparison with ground truth (in blue).}
    \label{fig:three_images}
\end{figure}

The data for training our pipeline is collected on a 3D curved surface. Then, we test the zero-shot accuracy on different planes, i.e., a planar surface and an inclined surface. The results can be seen in Table~\ref{tab:transferable}. However, by observing the output shape Fig~\ref{fig:three_images} b,c), the main part of the errors comes from the starting phase of estimation, and our model can capture the dynamics of power trajectory afterward. The mean percentage error increases compared to the test case. This error at the beginning phase could be due to the friction difference of varying planes. In the plane case, the robot accelerates faster, which causes an overestimation of the model. Meanwhile, for the inclined surface, the robot climbs the slope, causing underestimation. With a basic understanding of geometry in pre-planned trajectories, the performance of our pipeline can be further optimized by freezing the TCN layers and tuning only the parameters of output-dense layers.
\begin{table}[h]
\footnotesize
\begin{threeparttable}
\setlength\tabcolsep{1.5pt} 
\begin{tabular*}{\columnwidth}{@{\extracolsep{\fill}} l ccccccc}
\toprule
     Methods &
     \multicolumn{3}{c}{\textbf{Planar Surface}} & \multicolumn{3}{c}{\textbf{Inclined Surface}}  \\
\cmidrule{2-4}
\cmidrule{5-7}
     & MSE & MAPE(\%) &$\text{MAPE}_\text{sum}$(\%) & MSE & MAPE(\%) & $\text{MAPE}_\text{sum}$(\%)   \\
\midrule
     Expert's estimation & - & - & $51.017\pm25.17$ & - & - & $79.391\pm 6.82$ \\
     Our pipeline  & $0.070\pm 0.01$  & $14.981\pm 2.44$ &$11.567\pm 2.27$& $0.059\pm 0.03$ & $11.478\pm 7.05$  & $9.529\pm 7.96$\\
\bottomrule
\end{tabular*}
\vspace{5pt}
\caption{\textbf{Transferable Test in Zero-Shot Setup}: We test the zero-shot performance in unseen working surfaces and achieve considerable accuracy. It is worth mentioning that as shown in Fig.\ref{fig:three_images}, there is a certain amount of offset in the beginning phase, and the overall trend of the trajectory can be predicted, which means that the TCN layer in our pipeline can understand the geometry embedded in the trajectory information and adjust output accordingly. The error can be further minimized by partially training the output dense layer.}
\label{tab:transferable}
\end{threeparttable}
\end{table}

\begin{figure}[h]
    \centering
    \includegraphics[width=\linewidth]{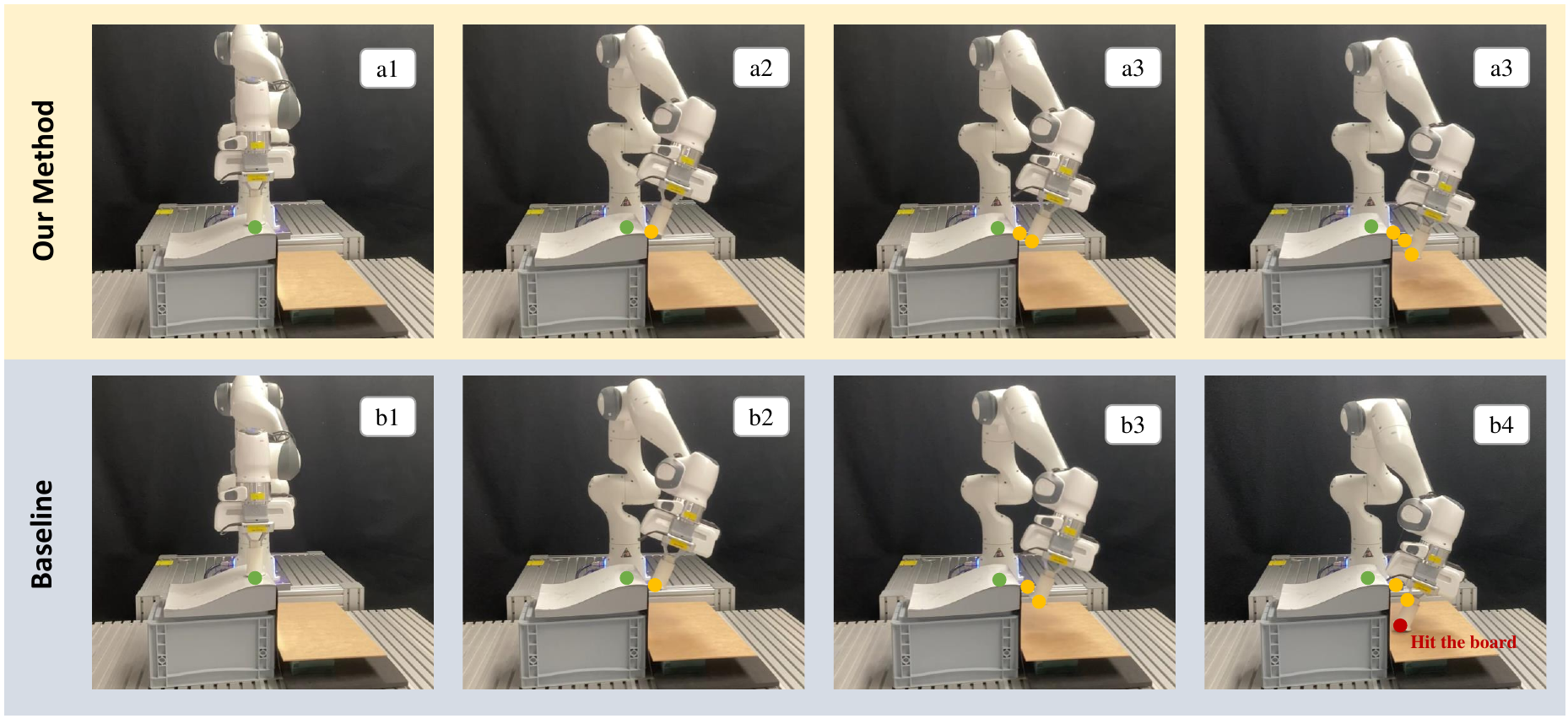} 
    \caption{\textbf{Safety Performance Comparison:} Our method and baseline performances are illustrated separately in the first and second rows. The green dots indicate maintained contact, the yellow dots indicate losing contact, and the red dots represent the worst-case scenario—hitting the board. In the first row, our method injects energy step-wise into the system. The energy within the robot system remains minimal and is frequently refreshed, preventing the robot from falling and hitting the working surface. In contrast, the baseline (shown in the second row) starts with a constant energy value. As a result, the controller cannot respond rapidly to the accident falling (losing contact), owing to the remaining energy in the energy tank. }
    \label{fig:fly}
\end{figure}

\subsection{Safety Performance Evaluation with Real Robot Setup} \label{exp:safe}
In this section, we highlight the improved safety performance of the robot system achieved through integrating our novel energy estimation method. As illustrated in Fig.~\ref{fig:fly}, we evaluate our pipeline with a tactile skill (trajectory from left to right and $f_\mathrm{d}=5N$) contact loss case that the beginning phase, where the baseline model provides scalar energy all at once and could potentially dangerous. In contrast to the baseline expert's estimation (detailed in \eqref{eq:dummyestimation}), our approach discretely injects the energy. It ensures the robot will only have the minimum required energy for the next time step, which significant safety enhancements in the whole process in the following aspects: i) Falling distance during contact loss and ii) Peak contact force.

\begin{table}[]
\centering
\begin{tabular}{lllll}
\cline{1-3}
\textbf{Methods}    & \textbf{Falling Distance/cm} & \textbf{Contact Force/N} & \textbf{} & \textbf{} \\ \cline{1-3}
Expert's estimation & $15.20\pm0$                   & $52.21\pm0.81$          &           &           \\
Our pipeline        & $9.78\pm0.38$               & no hitting               &           &           \\ \cline{1-3}
                    &                              &                          &           &          
\end{tabular}
\caption{Results corresponding to Fig.~\ref{fig:fly}}
\label{tab_last}
\end{table}
\section{Results and Discussion} \label{sec:results}
From the experiments, we can see that by leveraging our method, the robot can finish the task according to the estimated "just enough" energy trajectory, which shows better performance in terms of safety. Based on the experiment section, we demonstrate that nominal skill can be accurately morphed in the energy distribution landscape, ensuring the safety of pre-defined trajectory or even randomly generated trajectory without sacrificing performance.

\textbf{The Energy distribution matters in the contact rich tasks} As demonstrated in exp~\ref{exp:safe}, even with energy-based control, if the energy is given all at once as in the baseline case, the robot will take a much longer distance to stop or cause damage to the environment and the robot due to the exceeding energy in the system. Compared to the baseline estimation, in our method, the energy is discretely injected into the system following the estimated energy trajectory instead of given all energy all at once. This means that the robot will only have enough energy to go to the next step, improving the safety performance during the process, especially at the beginning phase. Our method makes the robot stop earlier than baseline and minimizes the contact force.

\textbf{Temporal Convolutional Network understands the geometry information embedded in the pre-planned trajectory and can be transferred to similar but different tasks} Thanks to the nature of the translation invariant of TCN, our pipeline can robustly estimate the results regardless of the relative position between working surface and the robot. Furthermore, even though all the data for training is collected on a curved surface, the pipeline can still produce a power trajectory that follows the shape of ground truth with reasonable errors apart from the beginning phase; details can be seen in experiment Section~\ref{transferable exp}. With this property and output layer fine-tuning, our pipeline enables transferable and reusable interaction skills between similar yet different surface geometry and material properties, including variations in friction. These acquired skills empower the robot to dynamically adapt its wiping and polishing technique based on encountered surface properties during diverse tasks.

\section{Conclusion and Limitations} \label{sec:conc}
We introduced the concept and mathematical approach of Tactile-Morph Skills, emphasizing its role in adaptive robotics with applications in tasks such as polishing and wiping. This study illustrates how externally provided nominal tactile skills can be made safe and performed well for an arbitrary desired motion and force profile adapted by the learned energy distribution of a surface. This approach allows for estimating the energy distribution of contact-rich trajectories on challenging surfaces, ensuring safe interaction without sacrificing performance. Moreover, this energy distribution knowledge can be transferred across different surface geometries with similar frictions, facilitating versatility in robotic applications.

Moreover, the demand for healthcare and home assistance automation is rising with the increasing demographic changes, such as an aging population. As the number of elderly individuals increases, there is a growing need for robots to assist with daily activities, provide companionship, and support medical care. By exploring the integration of energy-based control and data-driven learning, we aim to develop robust and adaptable robots that can revolutionize manufacturing and extend their applications to healthcare, home assistance, and beyond. This will ensure safety and performance in dynamic and complex environments, making intelligent robots more accessible and beneficial across multiple sectors.

However, the current trained model is limited by the size and variability of the training data, as the robot needs to be trained in the real world. Additionally, skill transfer between surfaces with different frictions is not viable with the current method. Thus, future work will focus on integrating surface friction into the learning module. This enhancement aims to improve the generalization capability of Tactile-Morph Skills, enabling more robust and adaptive robotic manipulation across a broader range of surface geometries and conditions.

\clearpage
\acknowledgments{We gratefully acknowledge the funding by the European Union’s Horizon 2020 research and innovation program as part of the project ReconCycle under grant no. 871352, the Bavarian State Ministry for Economic Affairs, Regional Development and Energy (StMWi) for the Lighthouse Initiative KI.FABRIK, (Phase 1: Infrastructure and the research and development program under grant no. DIK0249) CETI and Project Y.}


\bibliography{example}  
\newpage
\section{Supplementary Material} \label{sec:supp}
\subsection{\bf{Control Design}}
For an n-DOF robot manipulator under unified force-impedance control during contact with gravity compensation, the Lagrangian dynamics is
\begin{align}
\label{eq:dyn}
   \bm{M}(\bm{q})\ddot{\bm{q}}+\bm{C}(\bm{q},\dot{\bm{q}})\dot{\bm{q}}+\bm{g}(\bm{q}) &= \bm{\tau}_\mathrm{c}+\bm{\tau}_\mathrm{ext}\,,
   \\
     \bm{\tau}_\mathrm{c} &= \bm{\tau}_\mathrm{i} + \bm{\tau}_\mathrm{f} + \bm{\tau}_\mathrm{g}\,,
\end{align}
where $\bm{\tau}_\mathrm{ext} \in \mathbb{R}^n$ represents the external torque exerted on the robot, while $\bm{M}(\bm{q}) \in \mathbb{R}^\mathrm{n \times n}$ denotes the robot mass matrix, $\bm{C}(\bm{q}, \dot{\bm{q}})\dot{\bm{q}} \in \mathbb{R}^\mathrm{n}$ signifies the Coriolis and centrifugal vector, and $\bm{g} \in \mathbb{R}^\mathrm{n}$ stands for the gravity vector in joint space. Additionally, $\bm{\tau}_\mathrm{c} \in \mathbb{R}^\mathrm{n}$ represents the control torque applied by the robot, which encompasses the torque command for controlling motion and force explicitly and separately, with $\bm{\tau}_\mathrm{g} \in \mathbb{R}^\mathrm{n}$ representing gravity compensation. Moreover, $\bm{\tau}_\mathrm{i}$ and $\bm{\tau}_\mathrm{f} \in \mathbb{R}^\mathrm{n}$ denote torques individually introduced by impedance and force control, respectively. Subsequently, we develop a control algorithm for the input torque $\bm{\tau}_\mathrm{c}$ to execute the desired tactile manipulation skill. Unified force-impedance control governs the robot's response to external forces, ensuring compliance while following motion and force profiles separately and explicitly. Starting with the robot's dynamics equation in Cartesian space.
\begin{align}
\bm{M}_\mathrm{C}\ddot{\bm{x}} + \bm{C}_\mathrm{C}\dot{\bm{x}} +\bm{g}_\mathrm{C}&= \bm{f}_\mathrm{c}+\bm{f}_\mathrm{ext}\,,
\end{align}
where 
\begin{align}
\bm{M}_\mathrm{C} &= \bm{J}^\mathrm{\#T}\bm{M}\bm{J}^\mathrm{\#}\,,
\\
\bm{C}_\mathrm{C} &= \bm{J}^\mathrm{\#T}\bm{C}\bm{J}^\mathrm{\#}\,,
\\
\bm{g}_\mathrm{\mathrm{C}} &= \bm{J}^\mathrm{\#T}\bm{g}\,.
\end{align}
The external wrench to the base frame is denoted as $\bm{f}_\mathrm{ext} \in \mathbb{R}^6$. The robot mass matrix is represented as $\bm{M}_\mathrm{C}(\bm{q})$, where $\bm{q}$ is the joint configuration. The Coriolis and centrifugal effects are captured by $\bm{C}_\mathrm{C}(\bm{q}, \dot{\bm{q}}) \in \mathbb{R}^\mathrm{6 \times 6}$, and $\bm{g}_\mathrm{C}$ denotes the gravity vector in Cartesian space. Additionally, $\bm{f}_\mathrm{c}$ represents the wrench applied by the robot, which is related to the joint control torque $\bm{\tau}_\mathrm{c} \in \mathbb{R}^\mathrm{n}$ through the relationship $\bm{\tau}_\mathrm{c} = \bm{J}^\mathrm{T}(\bm{q}) \bm{f}_\mathrm{c}$, where $\bm{J} \in \mathbb{R}^\mathrm{6\times n}$ is the robot Jacobian matrix, and $\bm{J}^\#$ is the pseudo-inverse of the Jacobian. Compliance control, a subset of impedance control, omits inertia shaping and consequently excludes feedback of external forces. The compliance behavior is characterized by a stiffness matrix $\bm{K}_\mathrm{C} \in \mathbb{R}^\mathrm{6\times6}$ and damping behavior determined by a positive definite matrix $\bm{D}_\mathrm{C} \in \mathbb{R}^\mathrm{6\times6}$. Moreover, $\bm{x} \in \mathbb{R}^6$ denotes the current pose of the end-effector in the base frame, and the pose error is denoted by $\tilde{\bm{x}}$. A conventional compliance controller for motion tracking can be formulated as
\begin{align}
\tilde{\bm{x}} &= \bm{x}-\bm{x}_\mathrm{d}\,,
\\
    \bm{f}_\mathrm{i} &= -\bm{K}_\mathrm{C}\tilde{\bm{x}}-\bm{D}_\mathrm{C}\dot{\bm{x}}\,,
    \\
    \bm{\tau}_\mathrm{i} &= \bm{J}^T  \bm{f}_\mathrm{i}\,.
\end{align}

The force control is established to maintain the target contact force in the task space $\bm{f}^\mathrm{ee}_\mathrm{d} \in \mathbb{R}^6$, exerted by the robot concerning the external force $\bm{f}^\mathrm{ee}_\mathrm{\mathrm{ext}} \in \mathbb{R}^6$, as follows:
\begin{align}
    \bm{\tau}_\mathrm{f} &=  \bm{J}(\bm{q})^\mathrm{T} \bm{f}_\mathrm{f}, 
    \label{eq:frc} 
    \\
    \nonumber
    \bm{f}_\mathrm{f} &= \begin{bmatrix} [\bm{R}_\mathrm{ee}^\mathrm{0}]_\mathrm{3\times3} &  \bm{0}_\mathrm{3\times3} \\ \bm{0}_\mathrm{3 \times 3} & [\bm{R}_\mathrm{ee}^\mathrm{0}]_\mathrm{3\times3}
    \end{bmatrix} (\bm{f}_\mathrm{d}^\mathrm{ee} + \bm{K}_p \ \tilde{\bm{f}}_\mathrm{ext}^\mathrm{ee} + \\& \bm{K}_\mathrm{i} \int^t_0\tilde{\bm{f}}_\mathrm{ext}^\mathrm{ee} \ d\sigma)\, , \\
    \tilde{\bm{f}}^\mathrm{ee}_\mathrm{ext} &= \bm{f}^\mathrm{ee}_\mathrm{\mathrm{ext}}-\bm{f}^\mathrm{ee}_\mathrm{d}\, , 
\end{align}
In this context, $\bm{f}_\mathrm{f}$ $\in\mathbb{R}^6$ represents a feed-forward and feedback force controller in the base frame, which has been rotated by $\bm{R}_\mathrm{ee}^\mathrm{0}$. The proportional-integral (PI) controller gains are defined by the diagonal matrices $\bm{K}_\mathrm{p}$ and $\bm{K}_\mathrm{i}\in$ $\mathbb{R}^\mathrm{6\times6}$. The resultant control torque without the gravity compensation for unified force-impedance control $\bm{\tau} \in \mathbb{R}^n$ is
\begin{align}
        \bm{\tau} &=\bm{\tau}_\mathrm{f} + \bm{\tau}_\mathrm{i}\,.
\end{align}

\subsection{\bf{Network Architectures}}
This section provides a detailed description of our power estimation pipeline and its selected parameters.

\textbf{Temporal Convolutional Network:} We set up three layers of the TCN network for processing the time-variant inputs (desired trajectory $x_\mathrm{d}$, and desired velocity ${\dot{x}_\mathrm{d}}$~that differentiated from trajectory). Each TCN layer comprises a sequence of a dilated causal convolution, a weighted norm layer following a ReLU activation layer, a Dropout layer, and repeated once again. Besides the normal route, there is a residual route to perform an identical mapping, which is used at the last layer of the TCN network to output a power feature sequence. After power feature extraction, time-invariant input force policy $f_\mathrm{d}$ is concatenated to the end of power feature sequences and fed into the MLP power decoder.

\textbf{MLP Power Decoder:} To decode the power feature sequence extracted by TCN layers, we set up a power decoder with a 2-layer fully connected network.
\begin{table}[h]
\centering
\begin{tabular}{llllll}
\hline
\multicolumn{1}{l|}{} & \multicolumn{1}{l|}{Sliding Window} & \multicolumn{1}{l|}{kernel size} & \multicolumn{1}{l|}{nb stacks} & \multicolumn{1}{l|}{nb filters} & \multicolumn{1}{l}{dropout rate} \\ \hline
\multicolumn{1}{l|}{Pipeline} & \multicolumn{1}{l|}{100} & \multicolumn{1}{l|}{4} & \multicolumn{1}{l|}{1} & \multicolumn{1}{l|}{64} &\multicolumn{1}{l}{0.05}  \\ \hline 
\end{tabular}
\vspace{3pt}
\caption{\textbf{Parameters Summary.} }
\label{tab:parameters}
\end{table}

\subsection{\bf{Data Collection and Pre-processing}}
\begin{figure}[htbp]
    \centering
    \subfloat[\textbf{Data Collection and Experiment Setup.} All the data is collected with real robot execution on the 3D-printed white surface covered with a soft whiteboard to minimize the friction between the end-effector and the working surface.]{
        \includegraphics[width=0.45\textwidth]{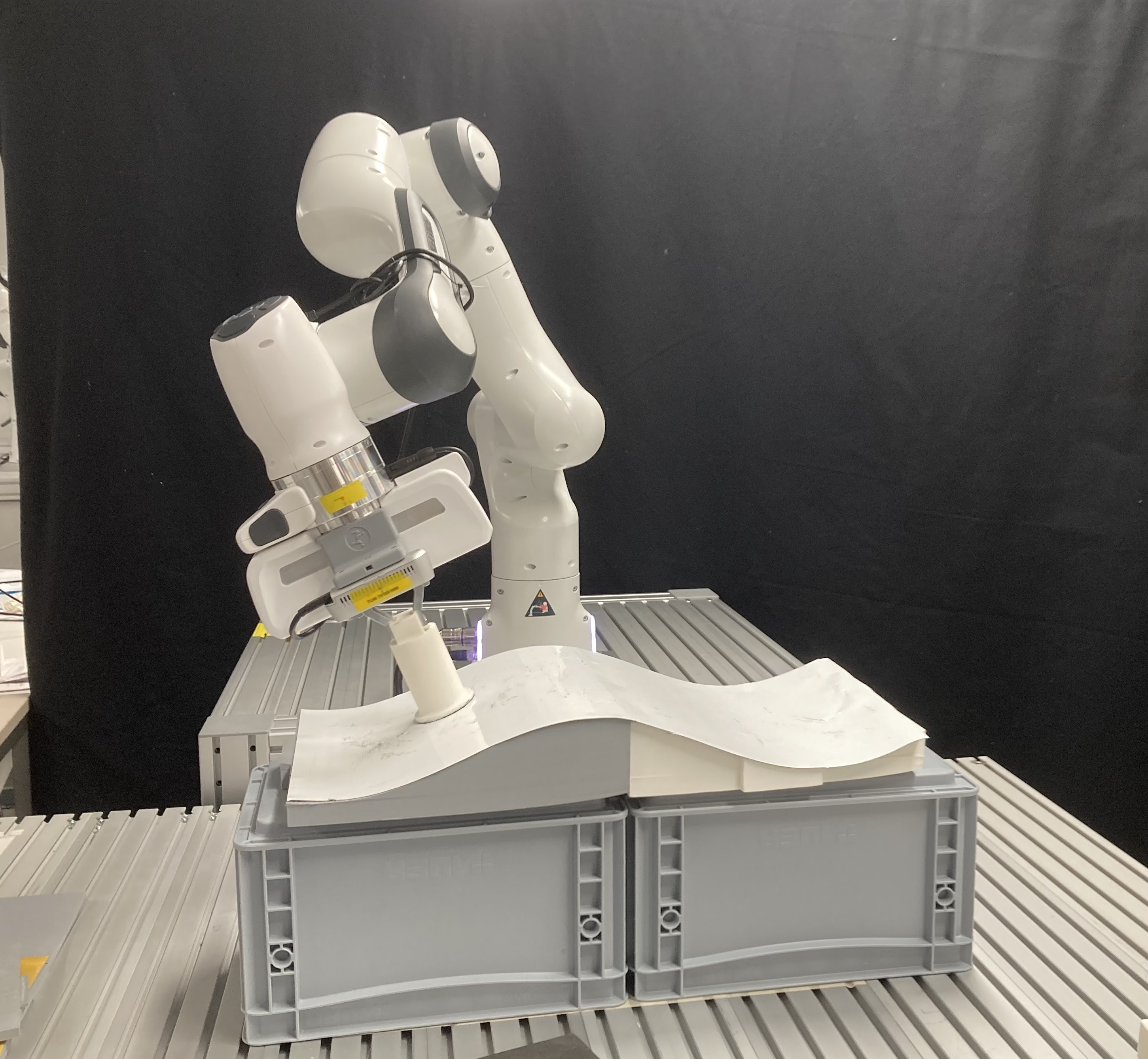}
        \label{fig:data_collection}
    }
    \hfill
    \subfloat[\textbf{Basic Motion Pattern.} We design four kinds of basic motion patterns for the data collection. The dot denotes the starting point, and the arrow is the endpoint. Row 4 is some randomly generated hand guide trajectory.]{
        \includegraphics[width=0.45\textwidth]{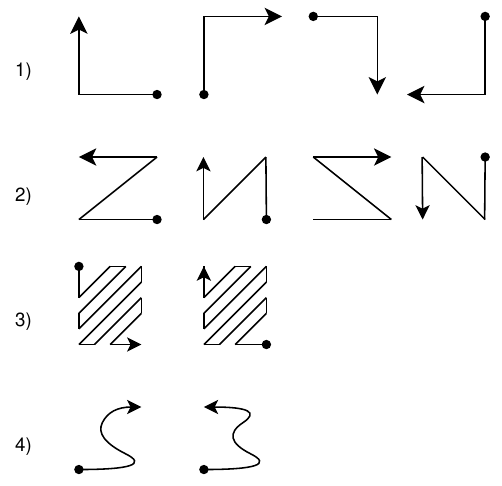}
        \label{fig:basic_pattern}
    }
    \caption{\textbf{Data collection Setup.}}
    \label{fig:combined}
\end{figure}

Due to the inherent complexity of contact and force closure between the robot and the environment, the data can't be easily gathered from simulation. Hence, we decided to collect data from a real robot wiping a 3d curve surface following basic motion patterns~\ref{fig:basic_pattern} to increase data accuracy and minimize the sim2real gap. All the data is collected on the working surface in Fig.\ref{fig:data_collection} in a 2-step manner. First, we pre-plan our trajectory based on a scan of the point cloud and save the planned trajectory into a CSV file. Then, we command the robot to read the CSV file and execute the pre-planned trajectory to gather ground truth power trajectory as a label value.

Due to the complexity of contact behavior, the collected data can be very noisy and not at the same scale. To bring all the data onto the same scale and improve the convergence speed, all the data will be normalized before training, and the power value of the data is applied with a $\log(y+3)$ transformation before normalization. 

\subsection{\bf{Training Details}}
Our pipeline is trained for roughly 2000 iterations (20 epochs). And we use the Adam optimizer with a learning rate of $10^{-4}$. The whole pipeline is trained using a mean absolute percentage error (MAPE)~\eqref{eq:mape}. Although the training process takes longer to converge, the overall performance in predicting the energy distribution for the given trajectory is better.
\begin{align}
\text { MAPE }=\frac{1}{n} \sum_{i=1}^n\left|\frac{y_i-\hat{y}_i}{y_i}\right|
\label{eq:mape}
\end{align}

\subsection{\bf{Experiment Details}}
The overall experiment setup includes several key components, as illustrated in Fig.\ref{fig:data_collection}. The primary equipment consists of a Franka robot controlled by a PC running Ubuntu 20.04 with a real-time kernel. The experiment involves three types of 3D-printed surfaces, each with a coefficient of friction $\mu \in [0.3, 0.4, 0.5]$. Additionally, a training PC equipped with an Intel i9-11900K CPU and Nvidia RTX 3090 GPU, also running Ubuntu 20.04, is utilized for the computations.


\subsection{Safety and Force Control Performance Evaluation with Real Robot Setup}
\begin{figure}
\centering
\includegraphics[width=\linewidth]{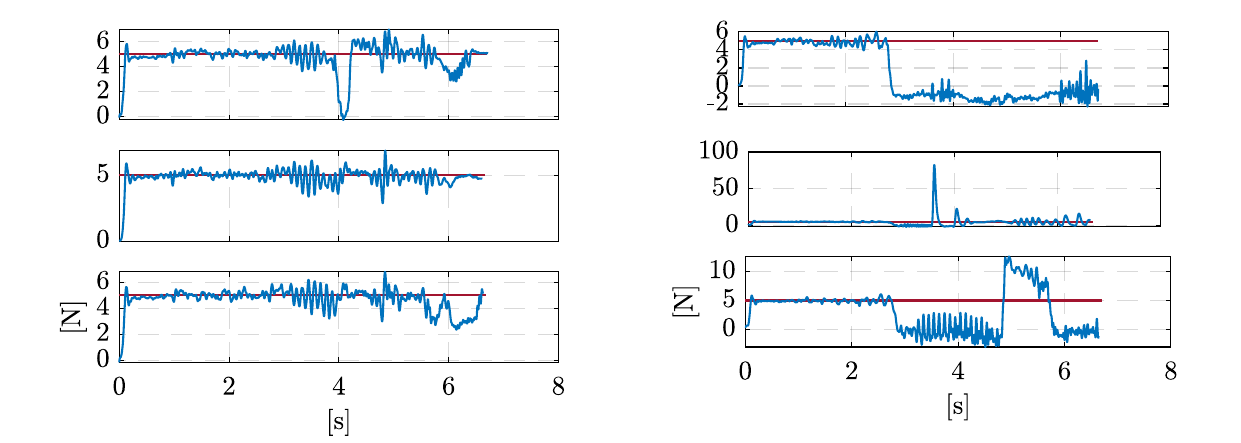}
\caption{\textbf{Force Controller Performance for Contact and Contact-Loss Scenarios Using Low, High, and Trained Task Energy.} Force tracking performance for the desired force of \SI{5}{N} during contact while following the desired trajectory a) for the pre-assigned low tank energy, b) high tank energy, c) trained energy distribution, force tracking performance during contact-loss while following the desired trajectory, d) for the pre-assigned low tank energy, e) high tank energy, f) trained energy distribution.}
\label{results:force}
\end{figure}

We compare the force tracking performance while following the desired motion for contact and contact loss scenarios on the curved surface: 1) with low tank energy (\SI{0.03}{J}), 2) with high tank energy (initiated with \SI{200}{J}), and 3) with carefully designed energy distribution from the estimation pipeline, as shown in Fig~\ref{results:force}. The robot system can accurately track force on a 3D surface with abundant and high tank energy; see Fig~\ref{results:force}b. However, storing high energy could potentially cause unsafe behavior during unexpected contact loss scenarios up to the impact force of \SI{100}{N}, as depicted in Fig~\ref{results:force}e. While the robot system with insufficiently low tank energy can finish the task safely, even if facing any, the force tracking performance is significantly worse than the high-energy case, as shown in Fig.~\ref{results:force}a. Executing the same task by predicting the required task energy for the given trajectory and force can preserve the accuracy of the high-energy case without sacrificing any performance, unlike the low-energy scenario. In contrast, the pre-assigned low-energy stops applying force faster than the high-energy case, as shown in Fig.~\ref{results:force}d and Fig.~\ref{results:force}e. This trade-off for performance and safety is handled nicely with our energy distribution estimation for the given trajectory and force profile, as seen in Fig.~\ref{results:force}c and Fig.~\ref{results:force}f. 

\end{document}